\documentstyle[aclap]{article}

\newcommand{\technical}[1]{\textsc{#1}}

\title{\vspace{-0.5in}Automatic Detection of Text Genre}
\author{Brett Kessler
\And
Geoffrey Nunberg
\And
Hinrich Sch\"utze
\AND
Xerox Palo Alto Research Center  \\
3333 Coyote Hill Road \\
Palo Alto CA 94304 USA \And
Department of Linguistics\\
Stanford University \\
Stanford CA 94305-2150 USA
\AND
email: {\tt \{bkessler,nunberg,schuetze\}@parc.xerox.com}\\
{\bf URL:} {\tt ftp://parcftp.xerox.com/pub/qca/papers/genre}
}

\begin{document}
\bibliographystyle{fullname}
\maketitle
\vspace{-0.5in}
\begin{abstract}
As the text databases available to users become larger and
more heterogeneous, genre becomes increasingly important for
computational linguistics as a complement to topical and structural
principles of classification. We propose a theory of genres as bundles
of {\em facets,} which correlate with various surface cues, and argue
that genre detection based on surface cues is as successful as
de\-tec\-tion based on deeper structural properties.
\end{abstract}

\section{Introduction}

Computational linguists have been concerned for the most part with two
aspects of texts: their \emph{structure} and their \emph{content}.
That is, we consider texts on the one hand as formal objects, and on
the other as symbols with semantic or referential values. In this
paper we want to consider texts from the point of view of genre; that
is, according to the various functional roles they play.

Genre is necessarily a heterogeneous classificatory principle, which
is based among other things on the way a text was created, the way it
is distributed, the register of language it uses, and
the kind of audience
it is addressed to. For all its complexity, this attribute can be
extremely important for many of the core problems that computational
linguists are concerned with. Parsing accuracy could be increased by
taking genre into account (for example, certain object-less
constructions occur only in recipes in English). Similarly for
POS-tagging (the frequency of uses of {\it trend} as a verb in the
{\it Journal of Commerce} is 35 times higher than in {\it Sociological
Abstracts}). In word-sense disambiguation, many senses are largely
restricted to texts of a particular style, such as colloquial or
formal (for example the word {\it pretty} is far more likely to have
the meaning ``rather'' in informal genres than in formal ones). In
information retrieval, genre classification could enable users to sort
search results according to their immediate interests. People who go
into a bookstore or library are not usually looking simply for
information about a particular topic, but rather have requirements of
genre as well: they are looking for scholarly articles about
hypnotism, novels about the French Revolution, editorials about the
supercollider, and so forth.

If genre classification is so useful, why hasn't it figured much in
computational linguistics before now? One important reason is that, up
to now, the digitized corpora and collections which are the subject of
much CL research have been for the most part generically homogeneous
(i.e., collections of scientific abstracts or newspaper articles,
encyclopedias, and so on), so that the problem of genre identification
could be set aside.  To a large extent, the problems of genre
classification don't become salient until we are confronted with large
and heterogeneous search domains like the World-Wide Web.

Another reason for the neglect of genre, though, is that it can be a
difficult notion to get a conceptual handle on, particularly in
contrast with properties of structure or topicality, which for all
their complications involve well-explored territory. In order to do
systematic work on automatic genre classification, by contrast, we
require the answers to some basic theoretical and methodological
questions. Is genre a single property or attribute that can be neatly
laid out in some hierarchical structure? Or are we really talking
about a multidimensional space of properties that have little more in
common than that they are more or less orthogonal to topicality? And
once we have the theoretical prerequisites in place, we have to ask
whether genre can be reliably identified by means of computationally
tractable cues.

In a broad sense, the word ``genre'' is merely a literary substitute for
``kind of text,'' and discussions of literary classification stretch back to
Aristotle. We will use the term ``genre'' here to refer to any widely
recognized
class of texts defined by some common communicative purpose or other
functional traits, provided the function is connected to some formal
cues or commonalities and that the class is extensible. For example an
editorial is a shortish prose
argument expressing an opinion on some matter of immediate public
concern, typically written in an impersonal and relatively formal
style in which the author is denoted by the pronoun {\it we}. But we
would probably not use the term ``genre'' to describe merely the class of
texts that have the objective of persuading someone to do something, since
that class --- which would include editorials, sermons, prayers,
advertisements, and so forth --- has no distinguishing formal properties.
At the other end of the scale, we would probably not use ``genre'' to
describe the class of sermons by John Donne, since that class, while it has
distinctive formal characteristics, is not extensible. Nothing hangs in the
balance on this definition, but it seems to accord reasonably well with
ordinary usage.

The traditional literature on genre is rich with classificatory
schemes and systems, some of which might in retrospect be analyzed as
simple attribute systems. (For general discussions of literary
theories of genre, see, e.g., \newcite{butcher32}, \newcite{dubrow82},
\newcite{fowler82}, Frye (1957)\nocite{frye57}, Hernadi
(1972)\nocite{hernadi72}, \newcite{hobbes08}, \newcite{staiger59}, and
\newcite{todorov78}.)  We will refer here to the attributes used in
classifying genres as \technical{generic facets.} A facet is simply a
property which distinguishes a class of texts that answers to certain
practical interests, and which is moreover associated with a
characteristic set of computable structural or linguistic properties,
whether categorical or statistical, which we will describe as
``generic cues.'' In principle, a given text can be described in terms
of an indefinitely large number of facets. For example, a newspaper
story about a Balkan peace initiative is an example of a
\technical{broadcast} as opposed to \technical{directed}
communication, a property that correlates formally with certain uses
of the pronoun {\it you}. It is also an example of a
\technical{narrative}, as opposed to a \technical{directive} (e.g., in
a manual), \technical{suasive} (as in an editorial), or
\technical{descriptive} (as in a market survey) communication; and
this facet correlates, among other things, with a high incidence of
preterite verb forms.

Apart from giving us a theoretical framework for understanding genres,
facets offer two practical advantages. First, some applications
benefit from categorization according to facet, not genre. For
example, in an information retrieval context, we will want to consider
the \technical{opinion} feature most highly when we are searching for
public reactions to the supercollider, where newspaper columns,
editorials, and letters to the editor will be of roughly equal
interest. For other purposes we will want to stress narrativity, for
example in looking for accounts of the storming of the Bastille in
either novels or histories.

Secondly, we can extend our classification to genres not previously
encountered. Suppose that we are presented with the unfamiliar
category \technical{financial analysts' report}. By analyzing genres
as bundles of facets, we can categorize this genre as
\technical{institutional} (because of the use of {\it we} as in
editorials and annual reports) and as \technical{non-suasive} or
non-argumentative (because of the low incidence of question marks,
among other things), whereas a system trained on genres as atomic
entities would not be able to make sense of an unfamiliar category.

\subsection{Previous Work on Genre Identification}

The first linguistic research on genre that uses quantitative methods
is that of Biber \shortcite{biber86,biber88,biber92,biber95}, which draws
on work on stylistic analysis, readability indexing, and differences
between spoken and written language. Biber ranks genres along several
textual ``dimensions'', which are constructed by applying factor
analysis to a set of linguistic syntactic and lexical features.  Those
dimensions are then characterized in terms such as ``informative vs.\
involved'' or ``narrative vs.\ non-narrative.'' Factors are not used
for genre classification (the values of a text on the various
dimensions are often not informative with respect to genre). Rather,
factors are used to validate hypotheses about the functions of various
linguistic features.

An important and more relevant set of experiments, which deserves
careful attention, is presented in \newcite{karlgren94}.  They too
begin with a corpus of hand-classified texts, the Brown corpus. One
difficulty here, however, is that it is not clear to what extent the
Brown corpus classification used in this work is relevant for
practical or theoretical purposes. For example, the category ``Popular
Lore'' contains an article by the decidedly highbrow Harold Rosenberg
from {\it Commentary,} and articles from {\it Model Railroader} and
{\it Gourmet,} surely not a natural class by any reasonable
standard. In addition, many of the text features in Karlgren and
Cutting are structural cues that require tagging. We will replace
these cues with two new classes of cues that are easily computable:
character-level cues and deviation cues.

\section{Identifying Genres: Generic Cues}\label{genericcues}

This section discusses generic cues, the ``observable'' properties of
a text that are associated with facets.

\subsection{Structural Cues}

Examples of structural cues are passives, nominalizations, topicalized
sentences, and counts of the frequency of syntactic categories (e.g.,
part-of-speech tags).  These cues are not much discussed in the
traditional literature on genre, but have come to the fore in recent
work \cite{biber95,karlgren94}.  For purposes of automatic
classification they have the limitation that they require tagged or
parsed texts.

\subsection{Lexical Cues}

Most facets are correlated with lexical cues.  Examples of
ones that we use are terms of address (e.g., \emph{Mr.}, \emph{Ms.}),
which predominate in papers like the \emph{New York Times}; Latinate
affixes, which signal certain highbrow registers like scientific
articles or scholarly works; and words used in expressing dates, which
are common in certain types of narrative such as news stories.

\subsection{Character-Level Cues}

Character-level cues are mainly punctuation cues and other separators
and delimiters used to mark text categories like phrases, clauses,
and sentences \cite{nunberg90}. Such features have not been
used in previous work on genre recognition, but we believe they have
an important role to play, being at once significant and very
frequent.  Examples include counts of question marks, exclamations
marks, capitalized and hyphenated words, and acronyms.

\subsection{Derivative Cues}

Derivative cues are ratios and variation measures derived from
measures of lexical and character-level features.

Ratios correlate in certain ways with genre, and have been widely used
in previous work. We represent ratios implicitly as sums of other cues
by transforming all counts into natural logarithms. For example,
instead of estimating separate weights $\alpha, \beta$, and $\gamma$
for the ratios words per sentence (average sentence length),
characters per word (average word length) and words per type
(token/type ratio), respectively, we express this desired weighting:
\[\alpha\log\frac{W+1}{S+1}+\beta
\log\frac{C+1}{W+1}+\gamma\log\frac{W+1}{T+1}\] as follows:
\begin{eqnarray*}
\lefteqn{(\alpha-\beta+\gamma)\log(W+1) -} \\
&&\alpha\log(S+1) + \beta\log(C+1) - \gamma \log(T+1)
\end{eqnarray*}
(where
W = word tokens, S = sentences, C = characters, T = word types). The
55 cues in our experiments can be combined to almost 3000 different ratios.
The log
representation
ensures that all these ratios are available implicitly while
avoiding overfitting and the high computational cost of training on a
large set of cues.

Variation measures capture the amount of variation of a certain count
cue in a text (e.g., the standard deviation in sentence length). This
type of useful metric has not been used in previous work on genre.

The experiments in this paper are based on 55 cues from the last three
groups: lexical, character-level and derivative cues. These cues
are easily computable in contrast to the structural cues that have
figured prominently in previous work on genre.

\section{Method}

\subsection{Corpus}

The corpus of texts used for this study was the Brown Corpus. For the
reasons mentioned above, we used our own classification system, and
eliminated texts that did not fall unequivocally into one of our
categories. We ended up using 499 of the 802 texts in the Brown
Corpus.  (While the Corpus contains 500
\emph{samples}, many of the samples contain several texts.)

For our experiments, we analyzed the texts in terms of three
categorical facets: \technical{Brow}, \technical{Narrative,} and
\technical{Genre}.  \technical{Brow} characterizes a text in terms of
the presumptions made with respect to the required intellectual
background of the target audience. Its levels are \technical{popular,
middle, uppermiddle,} and \technical{high}. For example, the
mainstream American press is classified as \technical{middle} and
tabloid newspapers as \technical{popular}. The \technical{Narrative}
facet is binary, telling whether a text is written in a narrative
mode, primarily relating a sequence of events. The \technical{Genre}
facet has the values \technical{reportage, editorial, scitech, legal,
nonfiction, fiction}. The first two characterize two types of articles
from the daily or weekly press: reportage and editorials. The level
\technical{scitech} denominates scientific or technical writings, and
\technical{legal} characterizes various types of writings about law
and government administration. Finally, \technical{nonfiction} is a
fairly diverse category encompassing most other types of expository
writing, and \technical{fiction} is used for works of fiction.

Our corpus of 499 texts was divided into a training subcorpus (402
texts) and an evaluation subcorpus (97). The evaluation subcorpus was
designed to have approximately equal numbers of all represented
combinations of facet levels. Most such combinations have six texts in
the evaluation corpus, but due to small numbers of some types of
texts, some extant combinations are underrepresented. Within this
stratified framework, texts were chosen by a pseudo random-number
generator. This setup results in different quantitative compositions
of training and evaluation set. For example, the most frequent genre
level in the training subcorpus is \technical{reportage}, but in the
evaluation subcorpus \technical{nonfiction} predominates.

\subsection{Logistic Regression}

We chose logistic regression (LR) as our basic numerical method. Two
informal pilot studies indicated that it gave better results than
linear discrimination and linear regression.

LR is a statistical technique for modeling a binary response variable
by a linear combination of one or more predictor variables, using a
logit link function:
\[g(\pi)=log(\pi/(1-\pi))\]
and modeling variance with a binomial random variable, i.e., the
dependent variable $log(\pi/(1-\pi))$ is modeled as a linear
combination of the independent variables. The model has the form
$g(\pi)=x_i\beta$ where $\pi$ is the estimated response probability
(in our case the probability of a particular facet value), $x_i$ is the feature
vector for text $i$, and $\beta$ is the weight vector which is
estimated from the matrix of feature vectors.  The optimal value of
$\beta$ is derived via maximum likelihood estimation \cite{McN}, using
SPlus \cite{splus91}.

For binary decisions, the application of LR was straightforward. For
the polytomous facets \technical{genre} and \technical{brow}, we
computed a predictor function independently for each level of each
facet and chose the category with the highest prediction.

The most discriminating of the 55 variables were selected using
stepwise backward selection based on the AIC criterion (see
documentation for \technical{step.glm} in \newcite{splus91}).  A separate
set of variables was selected for each binary discrimination task.

\subsubsection{Structural Cues}
In order to see whether our easily-computable surface cues are
comparable in power to the structural cues used in \newcite{karlgren94},
we also ran LR with the cues used in their experiment.  Because we
use individual texts in our experiments instead of the fixed-length
conglomerate samples of Karlgren and Cutting, we averaged all count
features over text length.

\subsection{Neural Networks}
Because of the high number of variables in our experiments, there is a
danger that overfitting occurs.  LR also forces us to simulate
polytomous decisions by a series of binary decisions, instead of
directly modeling a multinomial response.  Finally, classical LR does
not model variable interactions.

For these reasons, we ran a second set of experiments with neural
networks, which generally do well with a high number of variables
because they protect against overfitting.  Neural nets also naturally
model variable interactions. We used two architectures, a simple
perceptron (a two-layer feed-forward network with all input units
connected to all output units), and a multi-layer perceptron with all
input units connected to all units of the hidden layer, and all units
of the hidden layer connected to all output units.  For binary
decisions, such as determining whether or not a text is
\technical{Narrative}, the output layer consists of one sigmoidal
output unit; for polytomous decisions, it consists of four
(\technical{Brow}) or six (\technical{Genre}) softmax units (which
implement a multinomial response model) \cite{rdgc95}. The size of the
hidden layer was chosen to be three times as large as the size of the
output layer (3 units for binary decisions, 12 units for \technical{Brow}, 18
units for \technical{Genre}).

For binary decisions, the simple perceptron fits a logistic model just
as LR does. However, it is less prone to overfitting because we train
it using three-fold cross-validation. Variables are selected by
summing the cross-entropy error over the three validation sets and
eliminating the variable that if eliminated results in the lowest
cross-entropy error.  The elimination cycle is repeated until this
summed cross-entropy error starts increasing. Because this selection
technique is time-consuming, we only apply it to a subset of the
discriminations.

\section{Results}

Table~\ref{facetresults} gives the results of the experiments.  For each
genre facet, it compares our results using surface cues (both with
logistic regression and neural nets) against results using Karlgren
and Cutting's structural cues on the one hand (last pair of columns)
and against a baseline on the other (first column).  Each text
in the evaluation suite was tested for each facet.  Thus the
number 78 for \technical{Narrative} under method ``LR (Surf.) All''
means that when all texts were subjected to the \technical{Narrative}
test, 78\% of them were classified correctly.

There are at least two major ways of conceiving what the baseline should be
in this experiment.  If the machine were to guess randomly among $k$
categories, the probability of a correct guess would be $1/k$, i.e.,
$1/2$ for \technical{Narrative}, $1/6$ for \technical{Genre}, and
$1/4$ for \technical{Brow}.  But one could get dramatic improvement
just by building a machine that always guesses the most populated
category: \technical{nonfict} for \technical{Genre},
\technical{middle} for \technical{Brow}, and No for
\technical{Narrative}.  The first approach would be fair, because our
machines in fact have no prior knowledge of the distribution of genre
facets in the evaluation suite, but we decided to be conservative and
evaluate our methods against the latter baseline.  No matter which
approach one takes, however, each of the numbers in the table is
significant at $p < .05$ by a binomial distribution.  That is, there
is less than a 5\% chance that a machine guessing randomly could have
come up with results so much better than the baseline.

\begin{table*}
\caption{\label{facetresults}Classification Results for All Facets.}
\begin{center}

\begin{tabular}{l|c|rr|rr|rr|rr}
& Baseline & \multicolumn{2}{c|}{LR (Surf.)} 
& \multicolumn{2}{c|}{2LP} 
& \multicolumn{2}{c|}{3LP}
& \multicolumn{2}{c}{LR (Struct.)} \\
Facet          & & All & Sel. & All & Sel. & All & Sel. & All & Sel.\\\hline
Narrative &   54  &  78  &   80   & 82 & 82 &86 & 82 & 78  & 80\\
Genre     &  33   &  61  &   66   & 75 & 79 & 71 & 74 &66 &62\\
Brow      &  32   &   44  &   46   &47&---&54&---&46&53\\
\end{tabular}

\end{center}
Note.  Numbers are the percentage of the evaluation subcorpus ($N=97$) which
were correctly assigned to the appropriate facet level; the Baseline
column tells what percentage would be correct if the machine always
guessed the most frequent level.  LR is
Logistic Regression, over our surface cues (Surf.) or Karlgren and
Cutting's structural cues (Struct.); 2LP and 3LP are 2- or 3-layer
perceptrons using our surface cues.  Under each experiment, All tells
the results when all cues are used, and Sel.\ tells the results when
for each level one selects the most discriminating cues. A dash
indicates that an experiment was not run.
\end{table*} 

It will be recalled that in the LR models, the facets with more than
two levels were computed by means of binary decision machines for each
level, then choosing the level with the most positive score.
Therefore some feeling for the internal functioning of our algorithms
can be obtained by seeing what the performance is for each of these
binary machines, and for the sake of comparison this information is
also given for some of the neural net models.
Table~\ref{levelresults} shows how often each of the binary machines
correctly determined whether a text did or did not fall in a
particular facet level.  Here again the appropriate baseline could
be determined two ways.  In a machine that chooses randomly,
performance would be 50\%, and all of the numbers in the table would
be significantly better than chance ($p < .05$, binomial
distribution).  But a simple machine that always guesses No would
perform much better, and it is against this stricter standard that we
computed the baseline in Table~\ref{levelresults}.  Here, the binomial
distribution shows that some numbers are not significantly better than the baseline.  The
numbers that are significantly better than chance at $p<.05$ by the
binomial distribution are starred.

\begin{table*}
\caption{\label{levelresults}Classification Results for Each Facet Level.}
\begin{center}

\begin{tabular}{l|c|r@{}lr@{}l|r@{}l|r@{}l|r@{}lr@{}l}
& Baseline & \multicolumn{4}{c|}{LR (Surf.)} 
& \multicolumn{2}{c|}{2LP} 
& \multicolumn{2}{c|}{3LP}
& \multicolumn{4}{c}{LR (Struct.)} \\
Levels &        & 
\multicolumn{2}{c}{All} & \multicolumn{2}{c|}{Sel.} &
\multicolumn{2}{c|}{All} & 
\multicolumn{2}{c|}{All} & 
\multicolumn{2}{c}{All} & \multicolumn{2}{c}{Sel.}\\\hline
Genre &&&&&&&&&&&&&\\
\hspace{1em} Rep     &  81  &89&*&88&  &  94&* &94&*&90&* & 90&*\\
\hspace{1em} Edit    &  81  &75&&   75&  & 74&  &80&&79& & 77&\\
\hspace{1em} Legal   &  95  &96&&   96&  & 95&  &95&&93&&93&\\
\hspace{1em} Scitech &  94  &100&* &   96& & 99&*   &94&&93& &96&\\
\hspace{1em} Nonfict &  67  &67&  &   68&  & 78&*  &67&&73& &74&\\
\hspace{1em} Fict    &  81  &93&*&96&*  & 99&*  &81&&96&* & 96&*\\
Brow  &&&&&&&&&&&&&\\
\hspace{1em} Popular    &74 &74&  &75&    &74&&74&&72& & 73&\\
\hspace{1em} Middle     &68 &66&  &67&    &64&&64&&58& & 64&\\
\hspace{1em} Uppermiddle&88 &74&  &78&  &86&&88&&79& & 82&\\
\hspace{1em} High       &70 &84&* &88&* &89&* & 90&* &85&*&86&*\\
\end{tabular}

\end{center}
Note.  Numbers are the percentage of the evaluation subcorpus ($N=97$) which
was correctly classified on a binary discrimination task.  The Baseline
column tells what percentage would be got correct by guessing No for
each level.  Headers have the same meaning as in Table~\ref{facetresults}. 
\newline * means significantly better than Baseline at $p < .05$, using
a binomial distribution ($N$=97, $p$ as per first column).
\end{table*} 

Tables \ref{facetresults}~and~\ref{levelresults} present aggregate
results, when all texts are classified for each facet or level.
Table~\ref{confusion}, by contrast, shows which classifications are
assigned for texts that actually belong to a specific known level.
For example, the first row shows that of the 18 texts that really are
of the \technical{reportage Genre} level, 83\% were correctly classified as
\technical{reportage}, 6\% were misclassified as \technical{editorial},
and 11\% as \technical{nonfiction}.  Because of space constraints, we
present this amount of detail only for the six \technical{Genre}
levels, with logistic regression on selected surface variables.

\begin{table*}
\caption{\label{confusion}Genre Binary Level Classification Results
       by Genre Level.}
\begin{center}

\begin{tabular}{l|rrrrrr|r}
Actual & \multicolumn{6}{c}{Guess} \\
        &    Rep     &  Edit  &  Legal &Scitech&Nonfict&Fict    & N \\\hline
 Rep     &    83  &      6  &    0   &     0  &     11  &    0  & 18 \\
 Edit    &    17  &     61  &    0   &     0  &     17  &    6  & 18 \\
 Legal   &    20   &     0  &   20   &     0  &     60  &    0  & 5 \\
 Scitech &     0  &      0  &    0   &    83  &     17  &    0  & 6 \\
 Nonfict &     3  &     34  &    0   &     6  &     47  &    9  & 32 \\
 Fict    &     0  &      6  &    0   &     0  &      0  &   94  & 18 \\
\end{tabular}

\end{center}
Note.  Numbers are the percentage of the texts actually belonging to the
\technical{Genre} level indicated in the first column that were
classified as belonging to each of the \technical{Genre} levels
indicated in the column headers.  Thus the diagonals are correct
guesses, and each row would sum to 100\%, but for rounding error.
\end{table*}

\section{Discussion}

The experiments indicate that categorization decisions can be made
with reasonable accuracy on the basis of surface cues. All of the
facet level assignments are significantly better than a baseline of
always choosing the most frequent level (Table~\ref{facetresults}),
and the performance appears even better when one considers that the
machines do not actually know what the most frequent level is.

When one takes a closer look at the performance of the component
machines, it is clear that some facet levels are detected better than
others.  Table~\ref{levelresults} shows that within the facet
\technical{Genre}, our systems do a particularly good job on
\technical{reportage} and \technical{fiction}, trend correctly but not
necessarily significantly for \technical{scitech} and
\technical{nonfiction}, but perform less well for
\technical{editorial} and \technical{legal} texts.  We suspect that
the indifferent performance in \technical{scitech} and
\technical{legal} texts may simply reflect the fact that these genre
levels are fairly infrequent in the Brown corpus and hence in our
training set.  Table~\ref{confusion} sheds some light on the other
cases.  The lower performance on the \technical{editorial} and
\technical{nonfiction} tests stems mostly from misclassifying many
\technical{nonfiction} texts as \technical{editorial}.  Such confusion
suggests that these genre types are closely related to each other, as
in fact they are.  Editorials might best be treated in future
experiments as a subtype of \technical{nonfiction}, perhaps
distinguished by separate facets such as \technical{Opinion} and
\technical{Institutional Authorship}.

Although Table~\ref{facetresults} shows that our methods predict
\technical{Brow} at above-baseline levels, further analysis
(Table~\ref{levelresults}) indicates that most of this performance
comes from accuracy in deciding whether or not a text is
\technical{high Brow}. The other levels are identified at near baseline
performance. This suggests problems with the labeling of the
\technical{Brow} feature in the training data.  In particular, we had
labeled journalistic texts on the basis of the overall brow of the
host publication, a simplification that ignores variation among
authors and the practice of printing features from other publications.
We plan to improve those labelings in future experiments by
classifying brow on an article-by-article basis.

The experiments suggest that there is only a small difference between
surface and structural cues.  Comparing LR with surface cues and LR
with structural cues as input, we find that they yield about the same
performance: averages of 77.0\% (surface) vs.\ 77.5\% (structural) for
all variables and 78.4\% (surface) vs.\ 78.9\% (structural) for
selected variables.  Looking at the independent binary decisions on a
task-by-task basis, surface cues are worse in 10 cases and better in 8
cases.  Such a result is expected if we assume that either cue
representation is equally likely to do better than the other (assuming
a binomial model, the probability of getting this or a more extreme
result is $\sum_{i=0}^{8}b(i;18,0.5)=0.41$).  We conclude that there
is at best a marginal advantage to using structural cues, an advantage
that will not justify the additional computational cost in most cases.

Our goal in this paper has been to prepare the ground for using genre
in a wide variety of areas in natural language processing. The main
remaining technical challenge is to find an effective strategy for
variable selection in order to avoid overfitting during training. The
fact that the neural networks have a higher performance on average and
a much higher performance for some discriminations (though at the
price of higher variability of performance) indicates that
overfitting and variable interactions are important problems to
tackle.

On the theoretical side, we have developed a taxonomy of genres and
facets. Genres are considered to be generally reducible to bundles of
facets, though sometimes with some irreducible atomic residue. This
way of looking at the problem allows us to define the relationships
between different genres instead of regarding them as atomic entities.
We also have a framework for accommodating new genres as yet unseen
bundles of facets.  Finally, by decomposing genres into facets, we can
concentrate on whatever generic aspect is important in a particular
application (e.g., narrativity for one looking for accounts of the
storming of the Bastille).

Further practical tests of our theory will come in applications of
genre classification to tagging, summarization, and other tasks in
computational linguistics. We are particularly interested in
applications to information retrieval where users are often looking
for texts with particular, quite narrow generic properties:
authoritatively written documents, opinion pieces, scientific
articles, and so on. Sorting search results according to genre will
gain importance as the typical data base becomes increasingly
heterogeneous. We hope to show that the usefulness of retrieval tools
can be dramatically improved if genre is one of the selection criteria
that users can exploit.


\begin{thebibliography}{}

\bibitem[\protect\citename{Biber}1986]{biber86}
Biber, Douglas.
\newblock 1986.
\newblock Spoken and written textual dimensions in {E}nglish: {R}esolving the
  contradictory findings.
\newblock {\em Language}, 62(2):384--413.

\bibitem[\protect\citename{Biber}1988]{biber88}
Biber, Douglas.
\newblock 1988.
\newblock {\em Variation across Speech and Writing}.
\newblock Cambridge University Press, Cambridge, England.

\bibitem[\protect\citename{Biber}1992]{biber92}
Biber, Douglas.
\newblock 1992.
\newblock The multidimensional approach to linguistic analyses of genre
  variation: {A}n overview of methodology and finding.
\newblock {\em Computers in the Humanities}, 26(5--6):331--347.

\bibitem[\protect\citename{Biber}1995]{biber95}
Biber, Douglas.
\newblock 1995.
\newblock {\em Dimensions of Register Variation: A Cross-Linguistic
  Comparison}.
\newblock Cambridge University Press, Cambridge, England.

\bibitem[\protect\citename{Butcher}1932]{butcher32}
Butcher, S.~H., editor.
\newblock 1932.
\newblock {\em Aristotle's Theory of Poetry and Fine Arts, with The
  Poetics}.
\newblock Macmillan, London.
\newblock 4th edition.

\bibitem[\protect\citename{Dubrow}1982]{dubrow82}
Dubrow, Heather.
\newblock 1982.
\newblock {\em Genre}.
\newblock Methuen, London and New York.

\bibitem[\protect\citename{Fowler}1982]{fowler82}
Fowler, Alistair.
\newblock 1982.
\newblock {\em Kinds of Literature}.
\newblock Harvard University Press, Cambridge, Massachusetts.

\bibitem[\protect\citename{Frye}1957]{frye57}
Frye, Northrop.
\newblock 1957.
\newblock {\em The Anatomy of Criticism}.
\newblock Princeton University Press, Princeton, New Jersey.

\bibitem[\protect\citename{Hernadi}1972]{hernadi72}
Hernadi, Paul.
\newblock 1972.
\newblock {\em Beyond Genre}.
\newblock Cornell University Press, Ithaca, New York.

\bibitem[\protect\citename{Hobbes}1908]{hobbes08}
Hobbes, Thomas.
\newblock 1908.
\newblock The answer of mr {H}obbes to {S}ir {W}illiam {D}avenant's preface
  before {G}ondibert.
\newblock In J.E. Spigarn, editor, {\em Critical Essays of the Seventeenth
  Century}. The Clarendon Press, Oxford.

\bibitem[\protect\citename{Karlgren and Cutting}1994]{karlgren94}
Karlgren, Jussi and Douglass Cutting.
\newblock 1994.
\newblock Recognizing text genres with simple metrics using discriminant
  analysis.
\newblock In {\em Proceedings of Coling 94}, Kyoto.

\bibitem[\protect\citename{McCullagh and Nelder}1989]{McN}
McCullagh, P. and J.A. Nelder. 1989.
\newblock {\em Generalized Linear Models}, chapter~4, pages 101--123.
\newblock Chapman and Hall, 2nd edition.

\bibitem[\protect\citename{Nunberg}1990]{nunberg90}
Nunberg, Geoffrey.
\newblock 1990.
\newblock {\em The Linguistics of Punctuation}.
\newblock CSLI Publications, Stanford, California.

\bibitem[\protect\citename{Rumelhart \bgroup et al.\egroup }1995]{rdgc95}
Rumelhart, David~E., Richard Durbin, Richard Golden, and Yves Chauvin.
\newblock 1995.
\newblock Backpropagation: {T}he basic theory.
\newblock In Yves Chauvin and David~E. Rumelhart, editors, {\em
  Back-propagation: Theory, Architectures, and Applications}. Lawrence Erlbaum,
  Hillsdale, New Jersey, pages 1--34.

\bibitem[\protect\citename{Staiger}1959]{staiger59}
Staiger, Emil.
\newblock 1959.
\newblock {\em Grundbegriffe der Poetik}.
\newblock Atlantis, Zurich.

\bibitem[\protect\citename{Statistical Sciences}1991]{splus91}
Statistical Sciences.
\newblock 1991.
\newblock {\em {S-PLUS} Reference Manual}.
\newblock Statistical Sciences, Seattle, Washington.

\bibitem[\protect\citename{Todorov}1978]{todorov78}
Todorov, Tsvetan.
\newblock 1978.
\newblock {\em Les genres du discours}.
\newblock Seuil, Paris.

\end{thebibliography}
\end{document}